\documentclass[sigconf]{acmart}

\usepackage{graphicx}
\usepackage{booktabs}
\usepackage{multirow}
\usepackage{amsmath}
\usepackage{wrapfig}
\usepackage{enumitem}
\usepackage{xspace}

\usepackage[most]{tcolorbox}
\usepackage{xcolor}
\usepackage{colortbl}
\tcbset{before skip=6pt, after skip=6pt}

\AtBeginDocument{%
  }

\setcopyright{none} 
\copyrightyear{2026}
\acmYear{2026}
\acmDOI{XXXXXXX.XXXXXXX} 

\acmConference[Conference 'XX]{ACM Conference}{Month 2026}{Location}
\acmISBN{978-1-4503-XXXX-X/2018/06}

\begin{document}

\title{Towards Autonomous Memory Agents}

\author{Xinle Wu, Rui Zhang, Mustafa Anis Hussain, Yao Lu}
\affiliation{%
  \institution{National University of Singapore}
  \country{Singapore}
}

\renewcommand{\shortauthors}{Xinle Wu et al.}

\begin{abstract}
Recent memory agents improve LLMs by extracting experiences and conversation history into an external storage. This enables low-overhead context assembly and online memory update without expensive LLM training.  However, existing solutions remain \emph{passive and reactive}; memory growth is bounded by information that happens to be available, while memory agents seldom seek external inputs in uncertainties. 
We propose \emph{autonomous} memory agents that actively acquire, validate, and curate knowledge at a minimum cost. U-Mem materializes this idea via (i) a cost-aware knowledge-extraction cascade that escalates from cheap self/teacher signals to tool-verified research and, only when needed, expert feedback, and (ii) semantic-aware Thompson sampling to balance exploration and exploitation over memories and mitigate cold-start bias.
On both verifiable and non-verifiable benchmarks, U-Mem consistently beats prior memory baselines and can surpass RL-based optimization, improving HotpotQA (Qwen2.5-7B) by 14.6 points and AIME25 (Gemini-2.5-flash) by 7.33 points.
\end{abstract}

\keywords{Large Language Model, Memory, Agent, Thompson Sampling}


\maketitle

\section{Introduction}
Large language models (LLMs) nowadays empower AI agents to achieve higher and broader goals~\cite{liu2025advances,wang2024survey,fang2025comprehensive}, but deployed systems from research and industry are often ineffective in remembering user interactions, preferences, or domain knowledge beyond the given context window~\cite{hendrycks2025definition,gao2025survey,zhang2025survey}. This knowledge discontinuity forces AI agents to repeatedly ask known facts, past mistakes, and recompute answers; this clearly results in increased costs and degraded user experience~\cite{yin2025learning}. A recent way to address this problem to improve AI agents with experiences is to update the model parameters via supervised fine-tuning (SFT) or reinforcement learning (RL)~\cite{qi2024webrl,qian2025toolrl,li2025torl,luo2025agentmath}. However, doing so in practice is often too expensive and operationally infeasible~\cite{li2024revisiting}; besides, many foundation models are closed-source or difficult to update without regressions~\cite{comanici2025gemini,he2025evotest}. This motivates non-parametric solutions that can improve AI agents without touching the model weights~\cite{agrawal2025gepa,dong2024survey,wang2024agent,shinn2023reflexion}.

This has driven the recent interest in \emph{memory agents}~\cite{xu2025mem,chhikara2025mem0,ouyang2025reasoningbank,cao2025remember,zhang2026memrl}: instead of updating model weights, the agent extracts critical experiences from the conversation history, user preferences, task outcomes, domain knowledge, and reasoning patterns, and puts them into an external memory store~\cite{zhou2025simple,nan2025nemori,patel2025engram,yang2025learning,wei2025evo,tang2025agent,zhou2025memento}. Meanwhile, it retrieves and assembles relevant memories at query time to guide future decisions. This non-parametric, memory-based solution can be deployed on closed-source models with a low overhead and enables memory update when new experiences arrive~\cite{rasmussen2025zep,li2026just,huang2025licomemory,zhou2026wise,kong2025alphaopt,wang2025mirix}.

Despite this progress, existing memory agents remain largely \emph{passive}. Their memory growth and learning are typically bounded by whatever information happens to appear from user interaction. Some recent solutions~\cite{zhang2026memrl,ouyang2025reasoningbank,cao2025remember} additionally learn policies for memory operations for what to store, what to forget, and what to retrieve; this can even be done using RL-style objectives to optimize these choices~\cite{yan2025memory,ma2026fine,wang2025mem}. Yet, the memory operations are still reactive, where the agent does not actively seek external inputs when it encounters uncertainty, detects a failure mode, or identifies an opportunity to expand its knowledge base. In practice, there is abundant external data that can be harvested with minimal user frictions, including but not limited to user follow-up questions and corrections, more powerful teacher LLMs, tool-verified data and procedures, deep-research pipelines, and structured self-reflection over successes and failures. It is notable that exploiting these signals does not require human efforts in a traditional sense; instead, it can be framed as actions by the agent’s own autonomy in actively collecting, validating and making use of the available information.

\begin{figure}[t]
  \centering
    \includegraphics[width=1\linewidth]{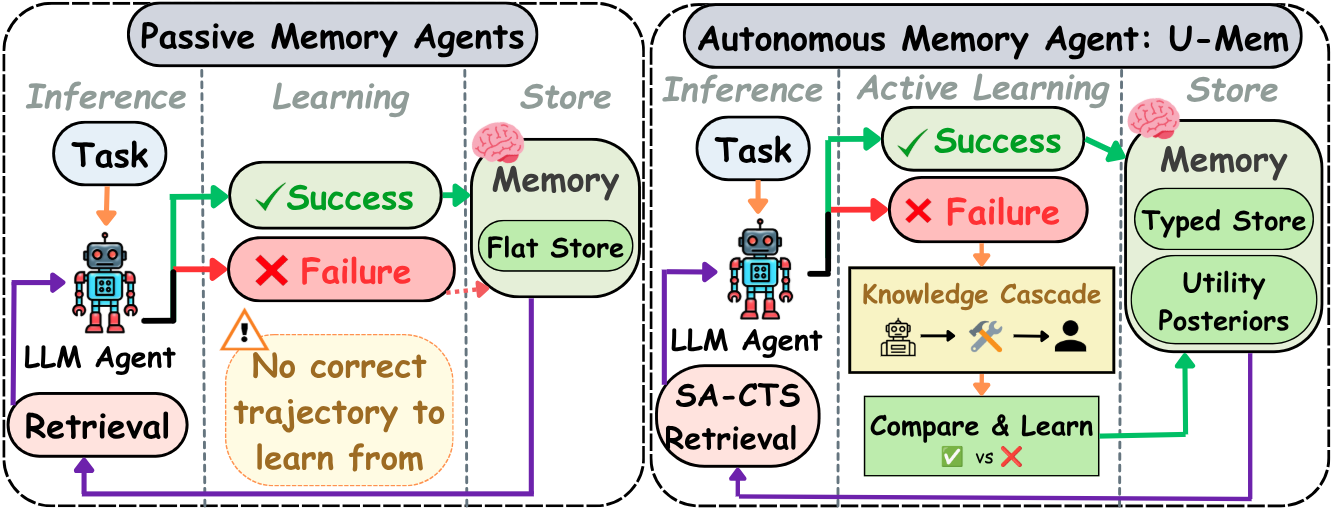}\vspace{-0.1in}
  \caption{Passive vs. autonomous memory agents.}
  \label{fig:intro}
\end{figure}

We argue that \emph{memory agents should be autonomous} in how they actively acquire, validate, and curate knowledge. As shown in Figure~\ref{fig:intro}, an autonomous memory agent goes beyond passive data logging and retrieval: it decides {when} additional information is needed, {where} to obtain it (e.g., from a teacher, tools, or research), and is aware of the cost that is going to incur. This clearly distinguishes from recent art in self-evolving agents that are mostly passive. An autonomous memory agent can still evolve over time without requiring constant human annotation; this enables personalization and domain adaptation in real applications. Nevertheless, autonomy does not preclude human supervision; rather, it treats {human experts as a last resort}, consulted only when cheaper knowledge sources fail to provide sufficient confidence. 


\begin{figure*}[t]
  \centering
    \includegraphics[width=0.7\linewidth]{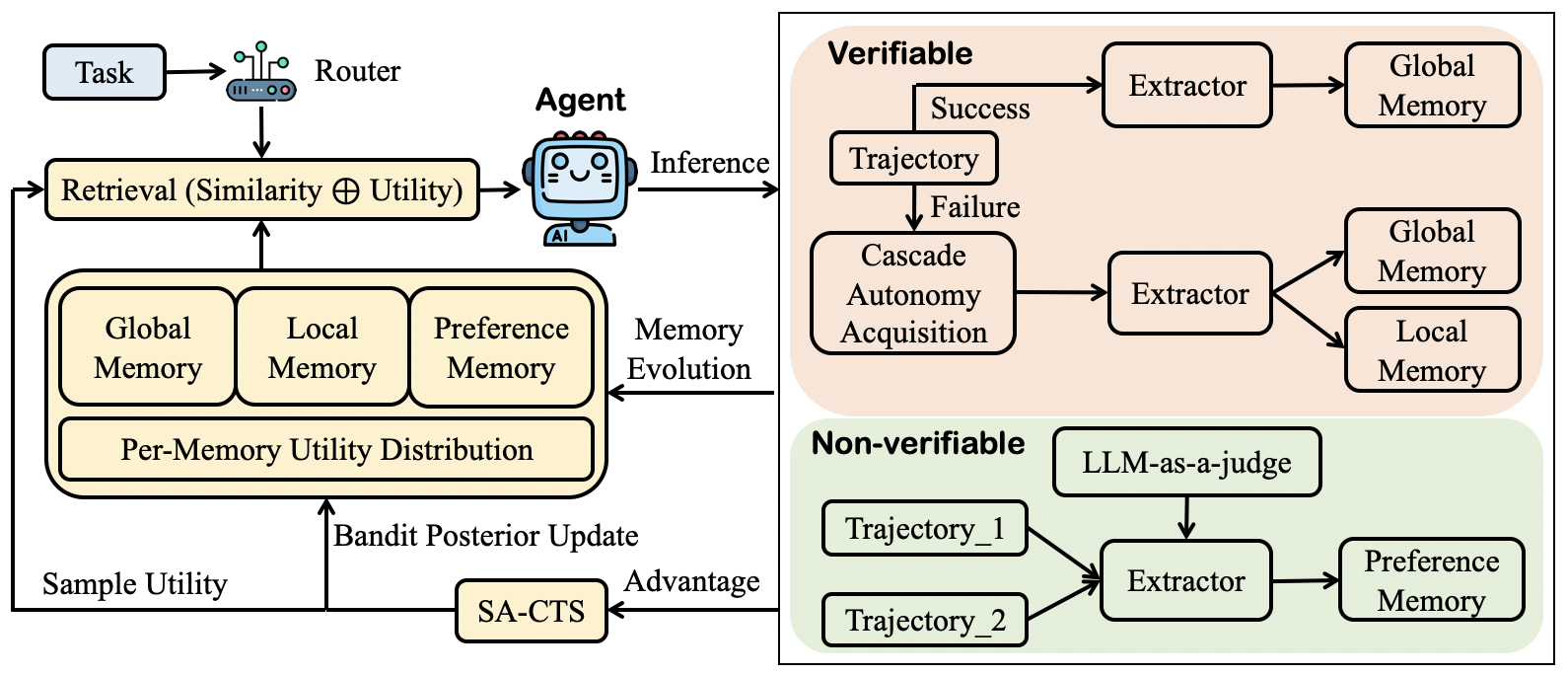}\vspace{-0.1in}
  \caption{Overview of U-Mem.
  }
  \label{fig:overview}
\end{figure*}

To realize this vision, we face new challenges. (1) \emph{Cost-awareness}. We note that, external knowledge, such as teacher LLMs, tool usage, research, is associated with different costs; the agent must decide who to query and when escalation is worth it. These issues are amplified in distribution shifts and cold-starts, where there is little feedback history to decide if new memories can be crucial for emerging needs. (2) \emph{Generalization}. Achieving high accuracy and generalization at the same time is non-trivial. The agent should provide a good level of accuracy across both \emph{verifiable} tasks, i.e., with objective checks, and \emph{non-verifiable tasks}, i.e., where correctness is ambiguous and feedback is model- or preference-related. 

In this paper, we take a preliminary step toward this direction and propose \emph{U-Mem}, the first {memory agent} that aims for autonomous knowledge acquisition and utility-driven update under cost constraints. To address the technical challenges, U-Mem leverages a cost-aware \emph{knowledge extraction cascade} that progressively harvests correct knowledge by escalating supervision only when necessary; the cascade starts from inexpensive self-reflection and more powerful teacher models, moves to tool-augmented verification and deep research when needed, and consults human experts as a last resort. To address the cold-start and adaptation issue, we develop \emph{semantic-aware Thompson sampling} to leverage feedback from downstream queries to improve the memory retrieval and utilization over time. By combining semantic relevance with posterior sampling over memory utility, the agent can systematically exploit proven memories while still exploring new ones.

To evaluate the effectiveness of autonomous memory agents, we conduct extensive experiments on diverse benchmarks with open-weight and proprietary models, spanning verifiable reasoning (e.g., AIME) and non-verifiable generation (e.g., AdvancedIF). 
Results demonstrate that \textsc{U-Mem} consistently outperforms state-of-the-art memory baselines and, remarkably, matches or exceeds expensive Reinforcement Learning (RL) baselines. 
For instance, \textsc{U-Mem} boosts the performance of Qwen2.5-7B on HotpotQA by over {14\%}, and further enhances strong proprietary models like Gemini-2.5-flash by {7.3\%} on AIME25. 
These findings validate the efficacy of autonomous memory acquisition and establish \textsc{U-Mem} as a robust, cost-effective method for evolving agents in dynamic deployment scenarios.

\vspace{0.05in}\noindent\textbf{Contributions.} We summarize our main contributions as follows:
\begin{itemize}[leftmargin=0.16in]
    \item Unlike memory agents that are often passive, we propose {autonomous memory agents} that proactively collect external knowledge and evidence to grow and refine long-term memory.
    \item We propose a {cost-aware knowledge extraction cascade} that turns both successes and failures into durable memory, adaptively escalating supervision (teacher, tools, deep research, human) to maximize correctness. We also develop {semantic-aware Thompson sampling} to balance exploration and exploitation and integrate semantic relevance with learned utility.
    \item We evaluate autonomous memory agents across both {verifiable} and {non-verifiable} tasks, demonstrating that autonomy can substantially extend the capability boundary of memory-based agent evolution under realistic budget constraints.
\end{itemize}

\section{Related Work}

\noindent\textbf{Reinforcement learning for LLMs}
is a standard paradigm in model fine-tuning and post-training alignment. RL training techniques have evolved from Proximal Policy Optimization (PPO)~\cite{schulman2017proximal} to more stable, direct techniques like DPO~\cite{rafailov2023direct} and the efficient, critic-free GRPO~\cite{guo2025deepseek}.
Methodology-wise, the training approaches diverge by feedback types: \textit{verifiable} domains (e.g., math) often utilize ground-truth outcome supervision (RLVR)~\cite{liu2024deepseek}, whereas \textit{non-verifiable} tasks often rely on preference models or AI feedback~\cite{lee2023rlaif}. While being effective, these paradigms remain computationally expensive due to the iterative parameter updates.

\vspace{0.05in}\noindent\textbf{LLMs with a memory bank}.
Existing research on LLM long-term memory can be categorized into two streams: dialogue memory, and reasoning trajectory memory.  
The former focuses on constructing a memory bank and extracting user profiles, preferences, and factual context from multi-session dialogues to enhance personalization and consistency.
For instance, {A-Mem}~\cite{xu2025mem} builds an agentic memory that stores long interactions as structured notes and continuously links and evolves them into an interconnected memory network for better long-term recall. {Mem0}~\cite{chhikara2025mem0} proposes a scalable memory pipeline that extracts, consolidates/updates, and retrieves salient information from multi-session dialogues to support efficient long-term context recall.
While effective for open-ended chat, these methods primarily address what the user said, rather than how to solve complex problems.

\vspace{0.05in}\noindent\textbf{Self-evolving memory agents}.
More relevant to our work is the second stream, which aims to distill transferable knowledge from the agent's reasoning trajectories. {ReasoningBank}~\cite{ouyang2025reasoningbank} leverages test-time scaling to sample multiple candidate paths, synthesizing memories by contrasting self-generated successful and failed trajectories.
{ReMe}~\cite{cao2025remember} proposes a framework for agent evolution that integrates multi-faceted experience distillation, context-adaptive reuse, and utility-based refinement.
A critical limitation of these approaches is their dependency on the LLM's intrinsic capability; they presume the model can eventually generate a correct solution via repeated sampling.
In essence, these systems remain {self-contained} and passive in knowledge harvesting; their knowledge growth is strictly limited to the model's pretraining data distribution. They are not designed to autonomously seek external knowledge when intrinsic reasoning fails, thus having "hard failures" where the model consistently makes errors. 


\vspace{0.05in}\noindent\textbf{Our ideas: autonomous memory agents}. Unlike off-the-shelf post-training and memory solutions that are often \textit{passive} and \textit{cost--agnostic}, \textsc{U-Mem} represents a shift towards \emph{autonomous} memory agents: (1) Our framework is \textit{active} to autonomously decide when to transcend internal limitations to acquire high-fidelity external knowledge from tools or even experts, and dynamically learns which memories are valuable through semantic-aware exploration. (2) On the other hand, our solution is also \textit{cost-aware} in both compute and data samples. Recent work such as {MemRL}~\cite{zhang2026memrl} and {Memento}~\cite{zhou2025memento} leverages query feedback to optimize the memory retrieval, but they overlook cold-start with new memories and various task difficulties. \textsc{U-Mem} aims to address these issues.

\section{Methods}


\subsection{Problem Formulation}
\label{subsec:problem_formulation}

We formalize the continuous improvement of an LLM agent as a streaming decision process. Consider a stream of tasks $\mathcal{Q} = \{q_1, q_2, \dots, q_T\}$ arriving sequentially.
The agent is parameterized by a frozen LLM backbone $\theta$ and a dynamic memory store $\mathcal{M}_t$.
At each time step $t$, the system state is defined as $S_t = \langle \theta, \mathcal{M}_t \rangle$. Unlike traditional Reinforcement Learning (RL) which updates $\theta$, our framework keeps $\theta$ fixed and evolves $\mathcal{M}_t$ to maximize performance.

The interaction workflow for a query $q_t$ proceeds as follows:
\begin{itemize}[leftmargin=0.2in]
    \item \emph{Retrieval:} The agent retrieves a subset of relevant memories $C_t = \mathcal{R}(q_t, \mathcal{M}_t)$ based on the current memory state.
    \item \emph{Inference:} Conditioned on the retrieved context, the frozen LLM generates a reasoning trajectory $y_t \sim \pi_\theta(\cdot | q_t, C_t)$.
    \item \emph{Feedback:} The environment provides a scalar feedback signal $r_t = \text{Eval}(y_t, q_t)$. We consider a unified feedback formulation covering both \textit{verifiable} tasks (where $r_t$ is a ground-truth binary outcome) and \textit{non-verifiable} tasks (where $r_t$ is a preference score from an external judge).
    \item \emph{Memory Evolution:} Based on the trajectory $y_t$ and feedback $r_t$, the memory extraction module $\mathcal{E}$ distills new insights to update the state: $\mathcal{M}_{t+1} \leftarrow \text{Update}(\mathcal{M}_t, \mathcal{E}(q_t, y_t, r_t))$.
\end{itemize}

\vspace{0.05in}\noindent\textbf{Objective.}
Our goal is to design an optimal retrieval policy $\mathcal{R}$ and extraction mechanism $\mathcal{E}$ that maximize the cumulative reward $\sum_{t=1}^T r_t$ over the task stream.
This formulation frames memory management as a dual problem to parameter optimization: maximizing long-term utility through experience accumulation rather than weight updates.

\vspace{0.05in}\noindent\textbf{Evaluation Protocol.}
To rigorously evaluate the "learning-by-using" capability, we adopt a \textit{Train-Test} protocol.
The agent first processes a \textit{Training Stream} where it is allowed to evolve $\mathcal{M}$ continuously.
Performance is then reported on a held-out \textit{Test Set}, where the memory state is frozen ($\mathcal{M}_{test} = \mathcal{M}_{final}$).

\subsection{Overview of \textsc{U-Mem}}
\label{subsec:framework_overview}

As illustrated in Figure~\ref{fig:overview}, U-Mem enables continuous, training-free evolution via a unified \textit{Retrieve-Infer-Evolve} cycle. The workflow commences with a task router (instantiated by the frozen LLM backbone) that directs incoming queries to verifiable or non-verifiable streams. In the \emph{Retrieval} phase, we employ Semantic-Aware Thompson Sampling (SA-CTS) to select memories by balancing semantic similarity with learned utility distributions. Subsequently, the agent performs \emph{Inference} using the enhanced prompt. Finally, during \emph{Memory Evolution}, a Cost-Aware Cascade (for verifiable tasks) or pairwise comparison (for non-verifiable tasks) distills experiences into memories, while environmental feedback updates memory utility posteriors to refine future retrieval without parameter updates.

\subsection{Memory for Verifiable Tasks}
\label{subsec:verifiable_memory}

In verifiable domains (e.g., mathematical reasoning, code generation), we posit that the substantial expansion of an agent's capability boundary stems from learning from failures rather than merely reinforcing successes. To this end, we introduce a {Cost-Aware Memory Extraction Cascade} to adaptively distill high-quality corrective insights from failure trajectories. For memory retrieval, we propose {Semantic-Aware Thompson Sampling (SA-CTS)}, a utility-driven retrieval mechanism designed to filter noise and precisely surface the most helpful memories for current reasoning.

\vspace{0.05in}\noindent\textbf{Memory Extraction Cascade}.
\label{subsubsec:cascade}
Standard memory approaches often prioritize extracting insights from successful trajectories. However, we argue that the most capability-expanding signals lie in failures.
Extracting memories from failures is non-trivial. Naive self-reflection on incorrect trajectories often leads to \textit{hallucinated attribution}, as the agent lacks the knowledge to identify its own errors. Recent work like ReasoningBank~\cite{ouyang2025reasoningbank} attempts to mitigate this via parallel sampling (generating $k$ trajectories). However, this fails when the model is fundamentally weak on a task, resulting in $k$ incorrect trajectories and no valid reference for learning.

To guarantee high-quality extraction, we introduce a \emph{Cost-Aware Cascade}. The core principle is that effective reflection requires a \textit{correct reference trajectory} ($\tau_{ref}$) for contrastive analysis. Upon generating a failure trajectory $\tau_{fail}$, we escalate the request for $\tau_{ref}$ through a hierarchy of increasing cost and reliability:

\begin{itemize}[leftmargin=0.2in]
    \item \emph{Level 1: Teacher LLM.} We first query a stronger model. If successful, its output serves as $\tau_{ref}$.
    \item \emph{Level 2: Tool-Augmented Teacher.} If the teacher model fails, we equip it with external tools (e.g., a code interpreter), which have been proven to significantly improve the correctness of LLMs.
    \item \emph{Level 3: Human Expert (Simulated).} As a last resort, we request a solution from a high-authority source, ideally a human expert. While expensive, this ensures the distillation of high-value, capability-expanding knowledge. In our experimental evaluation, to ensure scalability and reproducibility, we employ \texttt{Gemini-3-pro-preview} as a proxy for human experts.
    \end{itemize}

Once $\tau_{ref}$ is obtained, the agent performs {Contrastive Reflection} by analyzing the divergence between $\tau_{fail}$ and $\tau_{ref}$. The distilled insights are formalized into a structured memory schema $m = \langle \textit{Title}, \textit{Description}, \textit{Content} \rangle$ to support precise retrieval . We categorize these memories into two types:
\begin{itemize}[leftmargin=0.2in]
    \item \emph{Global Procedural Memory:} A high-level recipe summarizing the correct workflow derived from $\tau_{ref}$ (e.g., "For geometry proofs involving centroids, first apply the median property, then establish the segment ratios...").
    \item \emph{Local Corrective Memory:} Specific error-correction rules derived from the divergence point (e.g., "When calculating variance, do not square the sum directly; instead, calculate the mean of squares...").
\end{itemize}
If the agent succeeds initially, we bypass the cascade and directly extract Global Procedural Memory to reinforce the successful behavior.
We maintain separate stores for global and local memories, employing a flat storage structure.

\vspace{0.05in}\noindent\textbf{Semantic-Aware Thompson Sampling for Retrieval}.
\label{subsubsec:sa_cts}
Existing memory frameworks typically rely on dense vector retrieval based on semantic similarity. While effective for recall, this approach often retrieves "relevant noise"—memories that are semantically related to the query but offer no substantial aid to the reasoning process. Recent works like MemRL~\cite{zhang2026memrl} attempt to mitigate this by learning a scalar utility value for each memory based on environmental feedback (e.g., task success).

However, we identify two critical limitations in current utility-based retrieval paradigms such as MemRL:
\begin{itemize}[leftmargin=0.2in]
    \item \emph{Cold-Start Inequity (The Exploration Bias):} In streaming settings, newly created memories have not yet received feedback and thus lack reliable utility estimates (often undefined or conservatively initialized). Under deterministic greedy ranking, such memories are rarely retrieved, which prevents them from being evaluated and updated. This creates a self-reinforcing loop where older memories---having accumulated historical feedback---dominate retrieval, even when some new memories could be highly valuable.
    \item \emph{Reward Bias via Task Difficulty:} MemRL relies on absolute rewards as feedback. This introduces significant noise: a simple task may yield a positive reward even with irrelevant memories (false positive utility), while a complex task may result in failure despite retrieving useful memories (false negative utility). This confounds intrinsic task difficulty with memory efficacy.
\end{itemize}

To address these challenges, we propose \emph{Semantic-Aware Thompson Sampling (SA-CTS)}. Instead of maintaining a scalar point estimate, we model the utility of each memory $m_i$ as a Gaussian distribution $\mathcal{N}(\mu_i, \sigma_i^2)$, representing the estimated utility and the \textit{epistemic uncertainty}, respectively. This probabilistic formulation fundamentally resolves the exploration dilemma: even if a new memory's mean utility $\mu$ is not yet dominant, a large variance $\sigma^2$ allows it to be sampled at a high value. This grants it a chance to enter the top-$k$ context, receive feedback, and update its posterior—forming a closed-loop evolution mechanism absent in greedy approaches.

\vspace{0.05in}\noindent\emph{Semantic-Aware Initialization.}
To construct the initial prior for a new memory $m_{new}$, we transfer statistics from its $n$-nearest neighbors $\mathcal{N}_n(m_{new})$ in the embedding space. The distribution is initialized as:
\begin{equation}
    \mu_{new} = \frac{1}{n} \sum_{j=1}^n \mu_j, \quad \sigma_{new}^2 = \frac{1}{n} \sum_{j=1}^n \left( \sigma_j^2 + (\mu_j - \mu_{new})^2 \right) + \epsilon_{explore}
\end{equation}
Crucially, the hyperparameter $\epsilon_{explore}$ acts as a "hard exploration switch," guaranteeing a non-zero variance for all new memories to ensure meaningful exposure regardless of neighbor consensus.

\vspace{0.05in}\noindent\emph{Probabilistic Retrieval.}
During inference, given a query $q_t$, we employ Thompson Sampling to balance exploration and exploitation. For each candidate memory $m_i$, we sample a utility estimate $\tilde{u}_i$ from its posterior:
\begin{equation}
    \tilde{u}_i \sim \mathcal{N}(\mu_i, \sigma_i^2)
\end{equation}
The final retrieval score fuses semantic relevance with the sampled utility:
\begin{equation}
    \text{Score}(m_i) = (1-\lambda) \cdot \text{sim}(q_t, m_i) + \lambda \cdot \tilde{u}_i
\end{equation}
where $\lambda \in [0,1]$ balances the semantic prior and the learned utility. Unlike deterministic greedy selection, the stochasticity of $\tilde{u}_i$ allows high-variance (new or uncertain) memories to occasionally override low-variance established ones, dynamically optimizing the retrieval set.

\vspace{0.05in}\noindent\emph{Advantage-based Bayesian Update.}
To decouple memory efficacy from task difficulty (\textit{Reward Bias}), we employ an Advantage-based feedback mechanism. We define the signal as the performance differential between memory-augmented inference ($y_{mem}$) and base inference ($y_{base}$):
\begin{equation}
    r_{adv} = \mathcal{S}(y_{mem}) - \mathcal{S}(y_{base})
\end{equation}
For verifiable tasks, we instantiate $\mathcal{S}(y)$ as a binary correctness reward against ground truth, i.e., $\mathcal{S}(y)=\mathbb{I}[y \text{ is correct}] \in \{0,1\}$.
This differential reward isolates the marginal contribution of the retrieved memories, cancelling out the intrinsic difficulty of the task. 
Finally, we update the posterior of utilized memories via Bayesian approximation. Assuming likelihood noise variance $\sigma_{noise}^2$, the update rules are:
\begin{equation}
    \mu_{post} = \frac{\sigma_{prior}^2 r_{adv} + \sigma_{noise}^2 \mu_{prior}}{\sigma_{prior}^2 + \sigma_{noise}^2}, \quad \sigma_{post}^2 = \frac{\sigma_{prior}^2 \sigma_{noise}^2}{\sigma_{prior}^2 + \sigma_{noise}^2}
\end{equation}
This ensures the utility converges to the true marginal gain provided by the memory, robust to heterogeneous task difficulties.

\vspace{0.05in}\noindent\textbf{Streaming Memory Evolution and Maintenance}.
\label{subsubsec:maintenance}
In a continuous task stream, an append-only memory store inevitably suffers from unbounded growth and knowledge redundancy, diluting the retrieval density of high-value insights. To maintain a compact and evolving knowledge base, we implement an online Memory Consolidation mechanism.
Immediately following the extraction phase (Sec.~\ref{subsubsec:cascade}), the agent performs a semantic audit by comparing the newly generated memory $m_{new}$ against the retrieved context set $\mathcal{C}_{retrieved}$ used in the current task. Acting as a curator, the LLM executes one of three discrete operations to refine the store:

\begin{itemize}[leftmargin=0.16in]
    \item \emph{Append (Novelty):} If $m_{new}$ contains orthogonal insights not present in $\mathcal{C}_{retrieved}$, it is appended to the store, expanding the agent's knowledge boundary.
    \item \emph{Merge (Refinement):} If $m_{new}$ is semantically similar to an existing memory $m_{old}$ but offers superior clarity or generality, we replace $m_{old}$ with a merged version, consolidating fragmented knowledge into a stronger representation.
    \item \emph{Prune (Maintenance)}: If $m_{new}$ is a duplicate of an existing memory $m_{old}$ (or if $m_{old}$ is fully encompassed by a newly merged entry), the redundant $m_{old}$ is removed. This keeps the retrieval space compact and ensures high information density.
\end{itemize}

\subsection{Memory for Non-Verifiable Tasks}
\label{sec:non_verifiable}

In non-verifiable domains (e.g., creative writing, open-ended chat, or summarization), a definitive ground truth is absent, rendering the concept of binary ``success'' or ``failure'' inapplicable. However, we posit that the core principle of memory acquisition remains the same: \textit{learning from contrast}. Even without a binary signal, the agent can distill actionable insights by analyzing the divergence between a higher-quality response and a lower-quality one. To this end, we replace the \textit{Cost-Aware Cascade} used in verifiable tasks with a \emph{Pairwise Contrastive Extraction} strategy.

\vspace{0.05in}\noindent\textbf{Pairwise Contrastive Extraction}.
For a given task $q_t$, the agent generates two candidate reasoning trajectories. An external evaluator (e.g., an LLM-as-a-judge or human annotator) evaluates the pair to determine a winner, $y_{\text{win}}$, and a loser, $y_{\text{lose}}$.

The agent then acts as an extractor to analyze this pair and distill the preference into a structured template:
\begin{equation}
    \mathcal{M}_{\text{pref}} = \langle \text{\emph{Trigger}}, \text{\emph{Dimension}}, \text{\emph{Comparison}} \rangle
\end{equation}

\begin{itemize}[leftmargin=0.2in]
    \item \emph{Trigger:} The specific context or constraint where this preference applies (e.g., \textit{``When asked to write a concise marketing email...''}).
    \item \emph{Dimension:} The evaluation axis driving the decision (e.g., \textit{``Tone''}, \textit{``Clarity''}, or \textit{``Safety''}).
    \item \emph{Comparison:} An explicit, actionable directive derived from the contrast (e.g., \textit{``Prefer direct bullet points ($y_{\text{win}}$) over dense narrative paragraphs ($y_{\text{lose}}$)''}).
\end{itemize}

Relative signals are not only more robust to calibration noise but can also be used to induce a consistent preference signal in practice.
Finally, we directly reuse the {SA-CTS retrieval} and {streaming consolidation} mechanisms detailed in Sec. 3.3, where we define the advantage feedback as $r_{adv}=\mathbb{I}[y_{mem} \succ y_{base}]$, where $\succ$ denotes the LLM judge's pairwise preference.

\section{Experiments}
\label{sec:experiments}

\subsection{Experimental Setup}
\label{subsec:experimental_setup}


    

\noindent\textbf{Datasets and Protocols}.  
%
For \textit{verifiable tasks}, we employ \emph{AIME}~\cite{aops_aime_problems_and_solutions}, a challenging mathematics competition dataset, where we utilize historical problems (1983--2024) for knowledge acquisition and the held-out 2025 set for evaluation; and \emph{HotpotQA}~\cite{yang2018hotpotqa}, a multi-hop reasoning benchmark, where we use the training set for memory population and report performance on the official validation set. 
For \textit{non-verifiable tasks}, following~\cite{team2026kimi}, we select \emph{AdvancedIF}~\cite{he2025advancedif}, which evaluates the agent's ability to follow complex, multi-constraint instructions, and the STEM subset of \emph{HelpSteer3}~\cite{wang2025helpsteer3}, which assesses domain-specific helpfulness in science and engineering. Detailed statistics are provided in Table~\ref{tab:datasets}.
We provide our source code and data processing scripts in an anonymized repository\footnote{Available at: \url{https://anonymous.4open.science/r/code-release-456D/}}.

\begin{table}[t]
    \centering
    \caption{Datasets used in our experiments and their statistics.}
    \label{tab:datasets}
    \resizebox{\columnwidth}{!}{%
    \begin{tabular}{@{}l l l l@{}}
    \toprule
    \textbf{Dataset} & \textbf{Domain} & \textbf{Train Set} & \textbf{Test Set} \\
    \midrule
    AIME & Math Reasoning & 943 & 30 \\
    HotpotQA & Multi-hop QA & 2000 & 500 \\
    AdvancedIF & Instruction Following & 1316 & 329 \\
    HelpSteer3 (STEM) & Technical Helpfulness & 2000 & 500 \\
    \bottomrule
    \end{tabular}%
    }
\end{table}


\vspace{0.05in}\noindent\textbf{Baselines.}
We compare \textsc{U-Mem} against three categories of baselines: (1) \emph{No Memory}, the frozen LLM without parameter updates or external memory, serving as the performance lower bound; (2) State-of-the-art memory frameworks, including \emph{ReasoningBank}~\citep{ouyang2025reasoningbank}, which synthesizes memories from parallel trajectories, \emph{ReMe}~\citep{cao2025remember}, a dynamic procedural memory framework utilizing context-adaptive reuse, and \emph{MemRL}~\citep{zhang2026memrl}, which employs feedback-driven Q-learning to optimize retrieval utility; and (3) \emph{RL-Based Agents} representing the internalization paradigm, where we employ \emph{GRPO}~\citep{guo2025deepseek} for both verifiable tasks and non-verifiable tasks.

\vspace{0.05in}\noindent\textbf{Evaluation Metrics.}
For verifiable tasks, we report Accuracy on the test sets. 
For non-verifiable tasks, following~\cite{team2026kimi} we use an LLM-as-a-judge overall score with \texttt{gpt-5.1}. The judge outputs an aggregate rubric-based score for each response, and we report the mean normalized score (\%) on the test set.

\begin{table*}[t]
\centering
\caption{Main results on verifiable and non-verifiable tasks.}
\begin{tabular}{l|cc|cc|cc|cc}
\toprule
 & \multicolumn{4}{c|}{\textbf{Qwen2.5-7B}} & \multicolumn{4}{c}{\textbf{Qwen2.5-14B}} \\
\cmidrule(lr){2-5} \cmidrule(lr){6-9}
\textbf{Method} 
& \multicolumn{2}{c|}{Verifiable} & \multicolumn{2}{c|}{Non-verifiable}
& \multicolumn{2}{c|}{Verifiable} & \multicolumn{2}{c}{Non-verifiable} \\
\cmidrule(lr){2-3} \cmidrule(lr){4-5} \cmidrule(lr){6-7} \cmidrule(lr){8-9}
& AIME25 & HotpotQA & AdvancedIF & STEM & AIME25 & HotpotQA & AdvancedIF & STEM \\
\midrule
No Mem.        & $6.67 \pm 3.33$ & $37.80 \pm 2.17$ & $45.13 \pm 1.72$ & $56.16 \pm 2.72$ & $12.00 \pm 3.80$ & $60.00 \pm 2.74$ & $49.53 \pm 3.22$ & $62.56 \pm 3.01$ \\
ReasoningBank  & \underline{$9.33 \pm 1.49$} & $46.40 \pm 1.54$ & $44.98 \pm 0.57$ & $55.98 \pm 2.99$ & $12.67 \pm 2.79$ & $62.40 \pm 1.97$ & $49.01 \pm 2.72$ & $62.79 \pm 2.09$ \\
ReMe           & $8.00 \pm 1.83$ & $44.20 \pm 1.87$ & \underline{$47.05 \pm 1.32$} & \underline{$58.16 \pm 1.02$}  & $12.00 \pm 1.83$ & $61.80 \pm 2.43$ & $50.67 \pm 2.22$ & $63.98 \pm 1.88$ \\
MemRL          & $7.33 \pm 1.49$ & \underline{$47.80 \pm 2.16$} & $46.77 \pm 1.65$ & $57.37 \pm 1.78$ & $10.00 \pm 3.33$ & $64.40 \pm 2.17$ & $49.73 \pm 1.86$ & $64.03 \pm 2.14$ \\
RL-based       & $\boldsymbol{13.33 \pm 3.33}$ & $46.60 \pm 3.36$ & $46.50 \pm 1.48$ & $58.12 \pm 2.11$ & \underline{$17.33 \pm 2.79$} & \underline{$64.40 \pm 3.21$} & \underline{$51.02 \pm 2.71$}  & \underline{$64.72 \pm 2.31$} \\
\textbf{U-Mem}         & $\boldsymbol{13.33 \pm 2.36}$ & $\boldsymbol{52.40 \pm 0.89}$ & $\boldsymbol{49.34 \pm 0.86}$ & $\boldsymbol{58.68 \pm 1.72}$ & $\boldsymbol{18.67 \pm 1.83}$ & $\boldsymbol{65.20 \pm 1.79}$ & $\boldsymbol{51.28 \pm 1.64}$ & $\boldsymbol{65.99 \pm 0.93}$ \\
\bottomrule
\end{tabular}
\label{tab:main-results}
\end{table*}

\vspace{0.05in}\noindent\textbf{Implementation Details.}
We employ \emph{Qwen2.5-7B-Instruct} and \emph{Qwen2.5-14B-Instruct} as the primary backbone models to evaluate the scalability of our framework across different parameter scales. All experiments are conducted on a computational node equipped with $3 \times$ NVIDIA H200 GPUs.

\vspace{0.05in}\noindent\emph{RL Training Configurations (for Baselines).}
To construct competitive baselines, we implement Reinforcement Learning (RL) fine-tuning using Group Relative Policy Optimization (GRPO) with AdamW. We apply LoRA adapters with rank $r{=}16$, $\alpha{=}32$, and dropout set to $0.05$. Training is conducted with a learning rate of $1\mathrm{e}{-5}$ over 3 epochs, utilizing a question-grouped batch size of 4 and a gradient accumulation steps of 4. We enforce a KL regularization coefficient of $0.1$ and an advantage clipping threshold of $0.2$.
The reward design aligns with the task type:
(1) For \textit{verifiable tasks}, we use binary rewards ($r \in \{0, 1\}$) based on exact-match correctness against ground truth.
(2) For \textit{non-verifiable tasks}, we employ \texttt{Gemini-3.0-pro-preview} as an LLM-as-a-judge to assign rubric-based scores, which serve as the reward signal for GRPO optimization.

\vspace{0.05in}\noindent\emph{\textsc{U-Mem} Framework Configurations.}
For the autonomous memory mechanism, we configure the components as follows: (1) We utilize \texttt{Gemini-3.0-flash} as the Level-1 Teacher LLM. The Level-2 Tool-Augmented stage is supported by a secure Python code interpreter. For the Level-3 "Human Expert" proxy, we leverage \texttt{Gemini-3-pro-preview} to simulate high-quality, authoritative feedback. (2) We constrain the memory extraction module to distill a maximum of $m=3$ distinct insights per trajectory. (3) We employ \texttt{qwen3-4B-embedding} for vector representations. Retrieval is performed separately for each memory store: for verifiable tasks, we retrieve the top-$k=3$ items from both the Global Procedural and Local Corrective memory bases; for non-verifiable tasks, we retrieve top-$k=3$ from the Preference memory base. (4) We set the generation temperature to $\tau=0.7$ during the exploration/training phase and $\tau=0.2$ for final evaluation on the test set. (5) We set the exploration constant $\epsilon_{explore}=0.1$ and the likelihood noise variance $\sigma_{noise}^2=1.0$ for the Thompson Sampling mechanism. (6) We empirically set the fusion coefficient $\lambda=0.1$. (7) We use $n$=10 nearest neighbors for semantic initialization.
To ensure statistical reliability, all memory-based experimental results are reported as the mean and standard deviation averaged over 5 independent runs. Prompt templates can be found in Appendix~\ref{sec:prompts}.

\subsection{Main Results}
\label{subsec:main_results}

Table~\ref{tab:main-results} reports the comparative performance across verifiable and non-verifiable domains.
Key observations can be summarized as:
(1) \emph{Consistent State-of-the-Art Performance:} \textsc{U-Mem} achieves the best performance across all datasets and model scales, significantly outperforming the "No Memory" lower bound and existing memory baselines. While passive approaches like ReasoningBank and MemRL occasionally struggle to surpass the "No Memory" baseline (e.g., ReasoningBank on AdvancedIF with Qwen2.5-14B), \textsc{U-Mem} delivers robust improvements, demonstrating the efficacy of autonomous acquisition and distributional utility modeling (via SA-CTS).
(2) \emph{Competitive Alternative to RL:} Most notably, \textsc{U-Mem} matches or exceeds the performance of computationally expensive RL-based agents (GRPO) without modifying model parameters. On the challenging AIME25 mathematical reasoning task (Qwen2.5-14B), \textsc{U-Mem} outperforms the RL baseline ($18.67\%$ vs. $17.33\%$) with lower variance ($\pm 1.83$ vs. $\pm 2.79$), suggesting that actively curated memory can serve as a more stable and cost-effective substitute for continuous fine-tuning in dynamic deployment scenarios.

\subsection{Ablation Studies}

\begin{table}[t]
\centering
\caption{Component Ablations on HotpotQA (Qwen2.5-7B).}
\label{tab:ablations}
\begin{tabular}{l|cc}
\toprule
\textbf{Method} & \textbf{Accuracy (\%)} & \textbf{Expert Calls} \\
\midrule
\textit{Extraction Variants} & & \\
\quad Only Teacher (Level 1) & $49.40 \pm 1.45$ & 0\% \\
\quad Only Human Expert (Level 3) & $52.80 \pm 0.97$ & 100\% \\
\midrule
\textit{Retrieval \& Update Variants} & & \\
\quad w/o SA-CTS & $47.80 \pm 1.48$ & -- \\
\quad w/ MemRL's Retrieval & $50.40 \pm 0.96$ & -- \\
\quad w/o Streaming Update & $49.20 \pm 1.02$ & -- \\
\midrule
\textbf{\textsc{U-Mem} (Full Framework)} & \textbf{$52.40 \pm 0.89$} & \textbf{23.17\%} \\
\bottomrule
\end{tabular}
\end{table}

\noindent\textbf{Component Ablations}.
\label{subsubsec:ablations}
To validate the necessity of each component in \textsc{U-Mem}, we conduct an ablation study using the Qwen2.5-7B model on the HotpotQA dataset. We dissect the framework across three critical dimensions: extraction source (the cascade), retrieval strategy, and memory maintenance. The quantitative results are summarized in Table~\ref{tab:ablations}.

\vspace{0.05in}\noindent\emph{Efficacy of the Cost-Aware Cascade.}
We compare our adaptive cascade against two static baselines: \textit{Only Teacher}, which exclusively relies on the Level-1 model (lowest cost), and \textit{Only Human Expert}, which indiscriminately queries the Level-3 proxy for all failures (highest cost).
As shown in Table~\ref{tab:ablations}, \textit{Only Teacher} lags significantly behind \textsc{U-Mem} ($49.40\%$ vs. $52.40\%$). This performance gap highlights that while Level-1 is cost-effective, it hits a {capability ceiling} on complex reasoning tasks; access to stronger, authoritative supervision (i.e., the Expert) is indispensable for resolving the "hard tail" of failure cases that weaker models cannot self-correct.
Conversely, \textit{Only Human Expert} achieves the performance upper bound of $52.80\%$. Crucially, \textsc{U-Mem} retains {99.2\%} of this upper-bound performance while invoking the expensive expert for only {23.17\%} of the cases. This demonstrates the cascade's Pareto efficiency: it effectively filters out "shallow" errors solvable by weaker models, strategically reserving high-cost supervision only when essential to bridge the gap to peak performance. {For a comprehensive analysis of token consumption and training time overhead, please refer to Appendix~\ref{sec:cost_analysis}.}

\vspace{0.05in}\noindent\emph{Impact of SA-CTS.}
To isolate the contribution of our retrieval mechanism, we compare the full framework against two variants:
(1) \textit{w/o SA-CTS}: A vanilla RAG approach that retrieves memories solely based on static semantic similarity; and
(2) \textit{w/ MemRL's Retrieval}: A deterministic utility-based baseline that ranks memories using a greedy Q-value estimate.
Results in Table~\ref{tab:ablations} demonstrate the statistical superiority of \textsc{U-Mem}. 
First, it surpasses \textit{w/o SA-CTS} by a substantial margin of {4.6\%}, verifying that semantic relevance alone is insufficient—discerning \textit{functional utility} is essential to filter out "relevant but unhelpful" noise.
Second, \textsc{U-Mem} outperforms the state-of-the-art \textit{MemRL} mechanism by {2.0\%}. We attribute this gain to the exploration advantage of Thompson Sampling: unlike MemRL's greedy ranking which may permanently starve new, potentially high-value memories (the cold-start problem), SA-CTS's probabilistic sampling ensures fair exposure for new insights, enabling continuous discovery and effective self-evolution.

\vspace{0.05in}\noindent\emph{Necessity of Streaming Updates.}
The \textit{w/o Streaming Update} variant adopts an append-only strategy, disabling the merge and delete operations. This results in a performance degradation of {3.2\%}. We attribute this decline to \textit{memory saturation}: without active curation, the store becomes bloated with redundant or near-duplicate entries. These redundancies dilute the information density of the top-$k$ context window, crowding out diverse and orthogonal insights. This confirms that active memory consolidation—merging similar instances and pruning obsolete ones—is a prerequisite for sustaining long-term capability growth.

\vspace{0.05in}\noindent\textbf{Analysis of Learning Sources and Hyperparameters}.
\label{subsubsec:further_ablations}
To dissect the mechanisms driving \textsc{U-Mem}'s performance, we investigate two pivotal design choices: the specific value of learning from failures versus successes, and the sensitivity of the system to retrieval volume ($k$). The quantitative results are presented in Table~\ref{tab:source_and_k}.

\begin{table}[t]
\centering
\caption{Ablation studies on HotpotQA (Qwen2.5-7B): Impact of Learning Sources and Retrieval Hyperparameter $k$.}
\label{tab:source_and_k}
\begin{tabular}{l|c}
\toprule
\textbf{Setting} & \textbf{Accuracy (\%)} \\
\midrule
\multicolumn{2}{l}{\textit{Study A: Effectiveness of Learning Source}} \\
\quad Success Trajectories Only & $50.80 \pm 1.12$ \\
\quad Failure Trajectories Only (Contrastive) & $51.60 \pm 0.45$ \\
\quad \textbf{\textsc{U-Mem} (Hybrid)} & $\boldsymbol{52.40 \pm 0.89}$ \\
\midrule
\multicolumn{2}{l}{\textit{Study B: Effect of Retrieved Memory Number ($k$)}} \\
\quad $k=1$ & $50.20 \pm 1.10$ \\
\quad $k=3$ (Default) & $\boldsymbol{52.40 \pm 0.89}$ \\
\quad $k=5$ & $50.80 \pm 0.84$ \\
\quad $k=10$ & $48.80 \pm 0.84$ \\
\bottomrule
\end{tabular}
\end{table}

\vspace{0.05in}\noindent\emph{Efficacy of Learning from Failure.}
A core hypothesis of our work is that meaningful capability expansion stems primarily from correcting errors rather than merely reinforcing successes. To validate this, we compare the full hybrid framework against \textit{Success Only} and \textit{Failure Only} variants.
As shown in Table~\ref{tab:source_and_k} (Study A), \textit{Success Only} yields the lowest performance ($50.80\%$). We attribute this to the fact that success-based memories largely reinforce what the model already knows, providing limited marginal gain in capability.
Conversely, \textit{Failure Only} achieves a higher accuracy of $51.60\%$, confirming that contrastive reflection on errors effectively targets the agent's \emph{capability boundary}, rectifying specific "blind spots" and misconceptions.
However, the hybrid \textsc{U-Mem} outperforms both ($52.40\%$). This suggests a synergistic effect: failure-driven memories fix specific reasoning deficits (correction), while success-driven memories stabilize the execution of correct patterns on borderline tasks (consolidation), collectively forming a more robust policy.

\vspace{0.05in}\noindent\emph{Sensitivity to Retrieval Number ($k$).}
We analyze the impact of retrieval context size $k$ on inference accuracy, with results summarized in Table~\ref{tab:source_and_k} (Study B).
Performance initially improves as $k$ increases from 1 to 3 ($50.20\% \to 52.40\%$). This indicates that for multi-step reasoning tasks like HotpotQA, a single memory is often insufficient to cover the requisite evidential chain; a slightly larger context window improves informational recall.
However, performance degrades noticeably as $k$ extends to 5 and 10. We attribute this drop to context dilution and noise injection: despite the utility filtering of SA-CTS, retrieving a large number of memories inevitably introduces lower-relevance or distracting information into the prompt.
Empirically, $k=3$ strikes the optimal balance between providing sufficient guidance and maintaining a high signal-to-noise ratio.





\subsection{More Analysis}
\noindent\textbf{Scalability with Experience Accumulation}.
\label{subsubsec:scalability}
A critical requirement for any learning paradigm is \textit{scalability}: performance should improve predictably as the volume of available data increases. To verify whether \textsc{U-Mem} exhibits such scaling properties, we conduct an analysis on the HotpotQA dataset using Qwen2.5-7B. We vary the size of the source training set (experience pool) from 0 to 1,000 tasks, from which the agent autonomously extracts and populates its memory store.

\begin{figure}[htbp]
    \centering
    \includegraphics[width=0.8\linewidth]{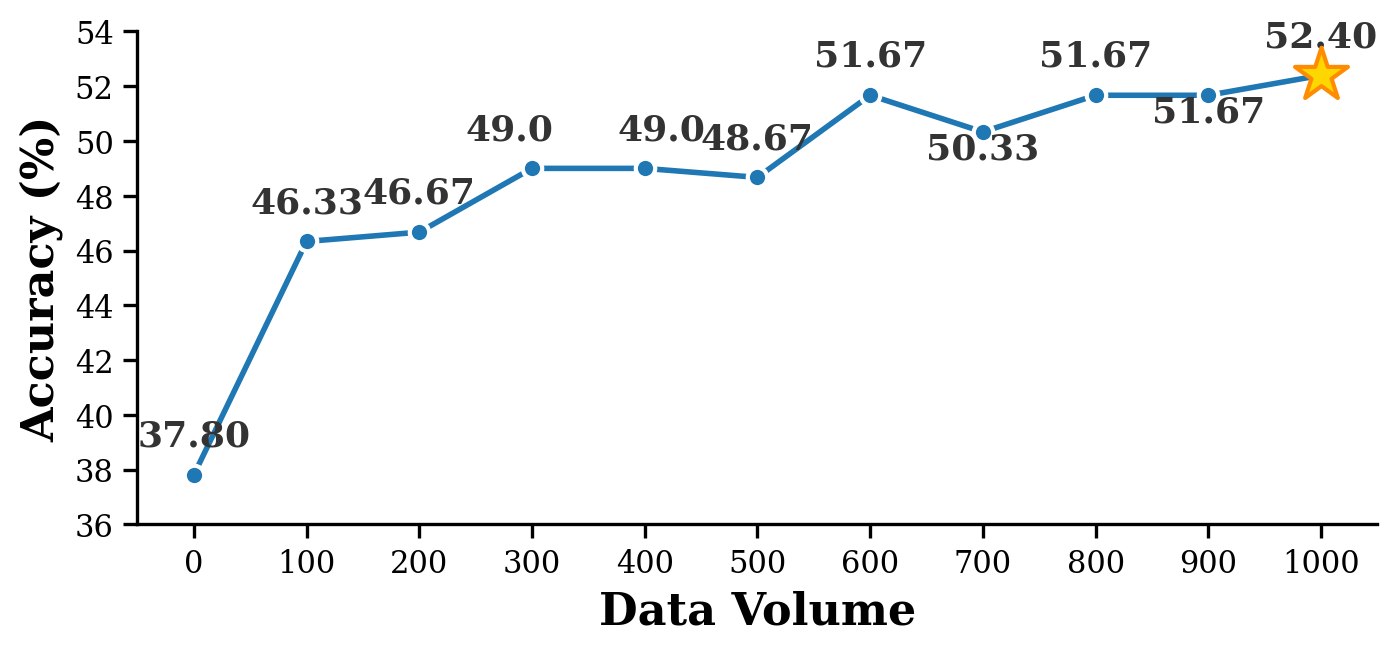}
    \caption{Scaling trend of \textsc{U-Mem} on HotpotQA.}    \label{fig:scaling_law}
\end{figure}

As illustrated in Figure~\ref{fig:scaling_law}, we observe a clear positive correlation between the scale of the experience pool and the agent's test accuracy. Starting from a baseline of $37.80\%$, the performance improves significantly as the agent "sees" more data, eventually peaking at $52.40\%$. Notably, the performance does not plateau or degrade significantly even as the memory pool grows larger. This indicates that \textsc{U-Mem}'s selection mechanism (SA-CTS) effectively manages the retrieval space, allowing the agent to continuously benefit from a larger scale of experience without being overwhelmed by noise. This confirms that \textsc{U-Mem} offers a scalable, data-driven pathway for agent evolution, akin to the scaling laws observed in pre-training but achieved non-parametrically.

\begin{table*}[htbp]
\centering
\caption{Performance comparison on stronger foundation models across four benchmarks.}
\begin{tabular}{l|cc|cc|cc|cc}
\toprule
 & \multicolumn{4}{c|}{\textbf{Gemini-2.5-flash}} & \multicolumn{4}{c}{\textbf{DeepSeek-chat}} \\
\cmidrule(lr){2-5} \cmidrule(lr){6-9}
\textbf{Method} 
& \multicolumn{2}{c|}{Verifiable} & \multicolumn{2}{c|}{Non-verifiable}
& \multicolumn{2}{c|}{Verifiable} & \multicolumn{2}{c}{Non-verifiable} \\
\cmidrule(lr){2-3} \cmidrule(lr){4-5} \cmidrule(lr){6-7} \cmidrule(lr){8-9}
& AIME25 & HotpotQA & AdvancedIF & STEM & AIME25 & HotpotQA & AdvancedIF & STEM \\
\midrule
No Mem.        & $46.67 \pm 4.08$ & $62.40 \pm 2.08$ & $58.62 \pm 2.41$ & $64.92 \pm 3.41$ & $46.67 \pm 2.36$ & $58.90 \pm 2.38$ & $54.82 \pm 2.49$ & $61.82 \pm 3.24$ \\
ReasoningBank  & \underline{$52.00 \pm 3.80$} & $68.40 \pm 2.77$ & $59.67 \pm 2.66$ & $64.99 \pm 2.10$ & \underline{$48.67 \pm 3.80$} & $60.80 \pm 2.14$ & \underline{$58.74 \pm 1.71$} & $62.22 \pm 1.48$ \\
ReMe           & $50.00 \pm 2.36$ & $69.20 \pm 1.08$ & \underline{$60.96 \pm 1.83$} & $65.02 \pm 2.84$ & $48.00 \pm 1.83$ & $59.80 \pm 2.06$ & $58.04 \pm 1.91$ & \underline{$63.88 \pm 1.73$} \\
MemRL & $50.67 \pm 2.79$ & \underline{$70.40 \pm 1.59$} & $59.83 \pm 2.37$ & \underline{$65.32 \pm 2.55$} & $48.00 \pm 1.83$ & \underline{$61.00\pm 1.83$} & $58.67 \pm 2.04$ & $63.12 \pm 0.95$ \\
\textbf{U-Mem}         & $\boldsymbol{54.00 \pm 2.79}$ &$\boldsymbol{71.60 \pm 1.09}$ & $\boldsymbol{62.87 \pm 0.94}$
 & $\boldsymbol{66.12 \pm 2.01}$ & $\boldsymbol{50.67 \pm 1.49}$ & $\boldsymbol{61.20 \pm 1.22}$ & $\boldsymbol{59.97 \pm 1.62}$ & $\boldsymbol{64.33 \pm 1.75}$ \\
\bottomrule
\end{tabular}
\label{tab:strong_models}
\end{table*}


\vspace{0.05in}\noindent\textbf{Impact of Task Distribution Similarity}.
\label{subsubsec:task_similarity}
We investigate the correlation between the similarity of the knowledge acquisition set to the evaluation set and the resulting performance gains. 
We quantify this alignment using the \emph{Average Maximum Cosine Similarity (AMCS)}, defined as:
\begin{equation}
    \mathcal{SS}(\mathcal{D}_{\text{test}}, \mathcal{D}_{\text{train}}) = \frac{1}{|\mathcal{D}_{\text{test}}|} \sum_{x \in \mathcal{D}_{\text{test}}} \max_{y \in \mathcal{D}_{\text{train}}} \text{sim}(\phi(x), \phi(y))
\end{equation}
where a higher $\mathcal{SS}$ indicates that for typical test queries, there exists a highly relevant counterpart in the history.

\begin{figure}[htbp]
  \centering
    \includegraphics[width=0.8\linewidth]{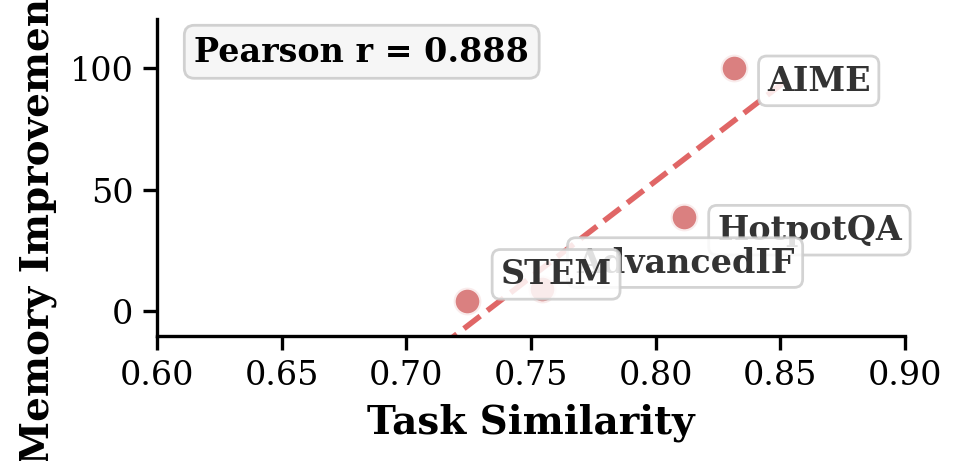}
  \caption{Correlation between task similarity and memory benefit. Each point denotes a task; the x-axis is AMCS (task similarity), and the y-axis is U-Mem’s performance gain over the base model. Pearson correlation: r = 0.888..
  }
  \label{fig:task_similarity}
\end{figure}

As illustrated in Figure~\ref{fig:task_similarity}, we observe a strong positive correlation (Pearson $r=0.888$) between the alignment score and the performance gains of \textsc{U-Mem} across benchmarks. Domains with high structural recurrence, such as AIME, exhibit the most significant improvements, whereas diverse tasks like STEM show more modest gains. These findings highlight the critical importance of \emph{embedding alignment and semantic overlap} in driving effective memory-based evolution, suggesting that non-parametric learning follows generalization laws similar to parametric methods.


\vspace{0.05in}\noindent\textbf{Scalability to Stronger Foundation Models}.
\label{subsubsec:strong_models}
While the previous comparisons utilized open-weight models (Qwen2.5 series) to enable a fair head-to-head comparison with parameter-update methods (RL), real-world agent deployments often rely on powerful proprietary or large-scale foundation models where gradient access is restricted.
To demonstrate the universality and scalability of \textsc{U-Mem}, we extend our evaluation to two strong models: {Gemini-2.5-flash} and {DeepSeek-chat}.
In this setting, RL-based baselines are omitted as parameter updates are operationally infeasible.

The results, summarized in Table~\ref{tab:strong_models}, demonstrate that \textsc{U-Mem} is model-agnostic and effective even atop highly capable baselines. We observe that \textsc{U-Mem} consistently achieves the best performance across all benchmarks, outperforming both the "No Memory" baseline and state-of-the-art memory frameworks (ReasoningBank, ReMe, and MemRL). This confirms its ability to further extend the capability boundary of even the strongest existing models.



\subsection{Case Study}

To visualize how \textsc{U-Mem} transcends the "internally bounded" limitations of passive memory agents, we present a qualitative analysis on a representative "hard failure" problem from AIME 2025. The task asks to find the number of subsets of 16 chairs chosen by 8 people such that "no person sits next to two other people," with the final answer modulo 1000.

The {No Memory} baseline fails consistently ($0/5$ success rate) due to a stubborn {inductive bias}: the model conflates the specific "no-three-in-a-row" constraint with the ubiquitous "no-two-adjacent" problem.
Existing memory agents struggle to correct this deep-seated reasoning error. {ReasoningBank} and \textit{ReMe}, relying on self-sampling, fail to generate a single correct trajectory to serve as a reference, resulting in zero effective memory extraction. \textit{MemRL}, while attempting to extract insights from the failure, suffers from {attribution error}: it extracts trivial procedural details—such as "calculating the number of gaps between people"—which are factually correct but insufficient to resolve the methodological flaw.
In contrast, \textsc{U-Mem} triggers the cost-aware cascade to consult a Level-1 Teacher LLM. By contrasting the model's failed trajectory with the successful external trajectory, \textsc{U-Mem} distills a high-order {Corrective Strategy}: (1) Explicitly define "Forbidden Patterns" to distinguish "no-three-in-a-row" from standard adjacency; and (2) When variable constraints involve upper bounds (e.g., a gap holding at most 2 people), immediately pivot from combinatorial formulas to Generating Functions. When re-evaluating the task, \textsc{U-Mem} retrieves this precise methodological correction, successfully overriding its intrinsic bias to derive the correct answer ($907$).

\section{Conclusion}

This paper presented U-Mem, a framework that transitions agents from passive information logging to {autonomous knowledge acquisition}. By synergizing a cost-aware cascade for {active acquisition} with Semantic-Aware Thompson Sampling (SA-CTS) for {adaptive retrieval}, U-Mem {represents a significant step towards autonomous memory agents}. We demonstrated that this non-parametric approach is consistently effective across verifiable and non-verifiable domains, achieving performance that matches or exceeds computationally expensive Reinforcement Learning (RL) methods. U-Mem thus establishes a scalable, data-efficient paradigm for continuous adaptation in dynamic environments.



\bibliographystyle{ACM-Reference-Format}
\bibliography{sample-base}

\appendix

\section{Reproducibility Statement}
To ensure the reproducibility of our work, we provide the source code in an anonymous GitHub repository available at: \url{https://anonymous.4open.science/r/code-release-456D/}. 
Regarding the datasets described in Table 1 and Section 4.1, we note that the partition between the training stream (experience pool) and the test set is generated dynamically. To guarantee the exact replication of the data splits used in our experiments, our repository includes the specific random seeds and preprocessing scripts required to reconstruct these partitions.

\section{Additional Experiments}

\subsection{Cost Efficiency Analysis}
\label{sec:cost_analysis}

\begin{figure}[htbp]
  \centering
    \includegraphics[width=0.9\linewidth]{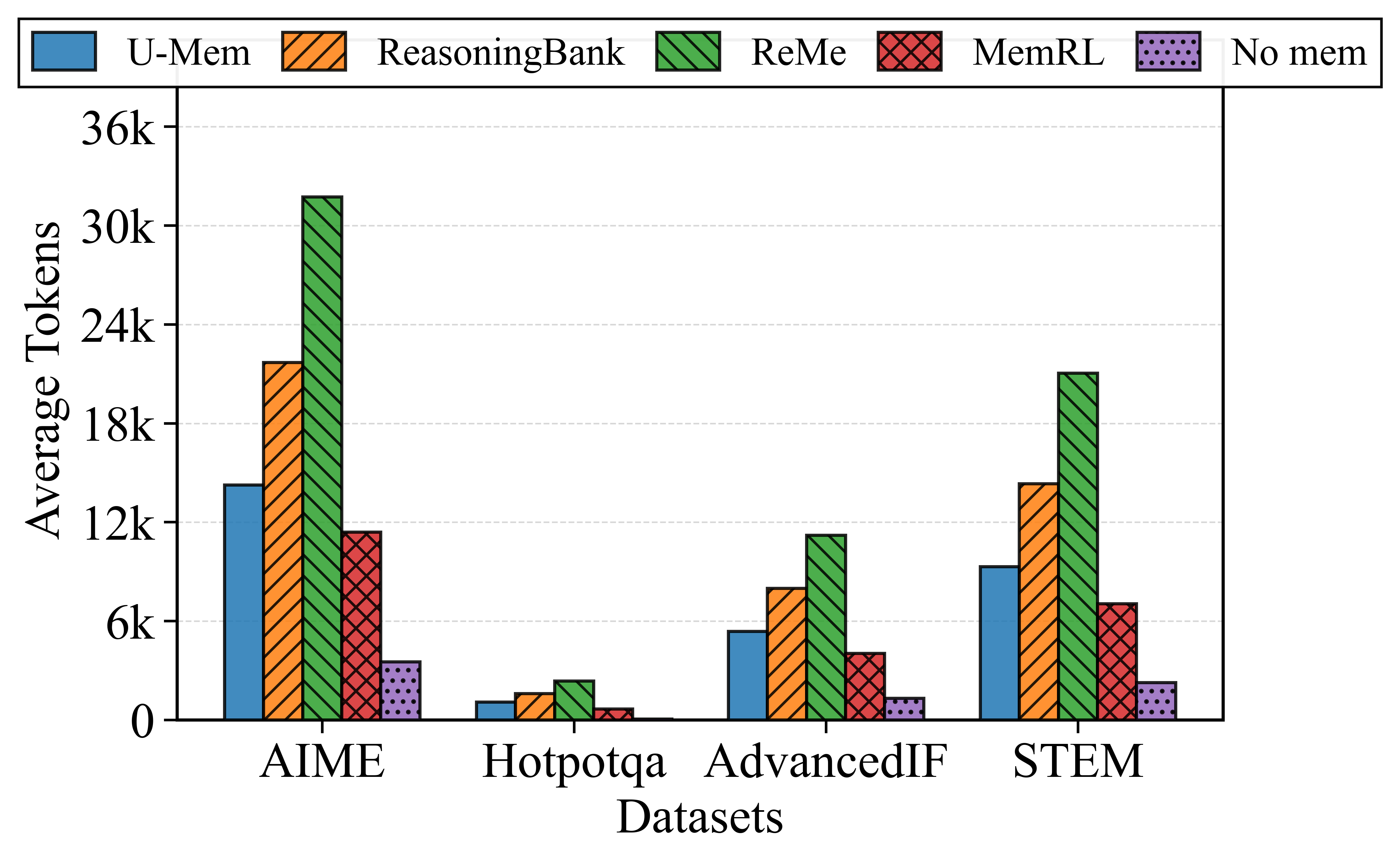}
  \caption{Comparison of average tokens usage between U-MEM, ReasoningBank, ReMe, MemRL and No Mem.}
  \label{fig:tokens}
\end{figure}

We evaluate the cost efficiency of U-MEM against baselines regarding token consumption (using \texttt{Gemini-2.5-flash}) and training time (using \texttt{Qwen2.5-7B}). For token consumption (averaged token number for each instance), Figure \ref{fig:tokens} shows that while all memory agents incur overhead compared to the {No Memory} baseline, U-MEM is significantly more efficient than sampling-heavy methods like ReasoningBank and ReMe, which require processing parallel trajectories. Although U-MEM consumes slightly more tokens than MemRL, this investment is justified by its superior performance, avoiding the attribution errors common in MemRL's low-cost but shallow reflection.

\begin{figure}[htbp]
  \centering
    \includegraphics[width=0.9\linewidth]{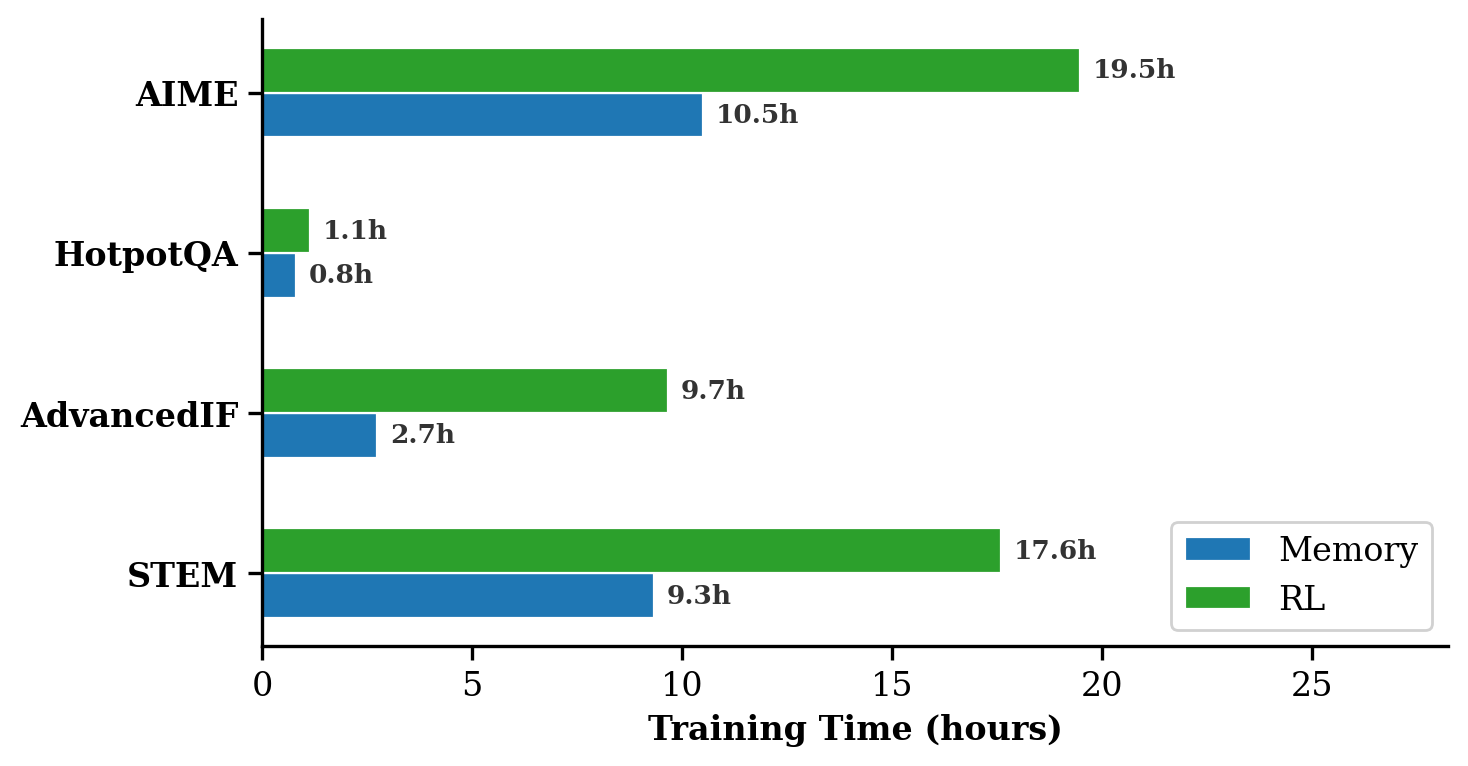}
  \caption{Comparison of training time (GPU hours) between U-MEM and RL (GRPO). U-MEM achieves comparable or better performance with significantly reduced time costs.}
  \label{fig:training_time}
\end{figure}

Regarding training time (which, for U-MEM, denotes the streaming memory acquisition phase without parameter updates), Figure \ref{fig:training_time} demonstrates that U-MEM consistently outperforms the RL baseline (GRPO). On the AIME benchmark, U-MEM completes memory acquisition in just 10.5 hours compared to 19.5 hours for RL fine-tuning—a near $2\times$ speedup. Similar efficiency gains are observed on AdvancedIF (2.7h vs. 9.7h) and STEM (9.3h vs. 17.6h). These results highlight U-MEM as a highly efficient, gradient-free alternative to RL, capable of continuous evolution with significantly lower computational latency.

\section{Prompts}
\label{sec:prompts}

We provide example prompts used in \textsc{U-Mem}. 
First, the prompts for {Memory Extraction}—covering both success summarization and failure correction—are presented in Figures~\ref{AIME_success_prompt}, \ref{AIME_prompt}, and \ref{fig:nonverifiable-pairwise-memory-extraction}.
Second, the prompts for {Memory-Augmented Inference}, which integrate retrieved insights into the reasoning context, are shown in Figures~\ref{fig:aime-solve-with-memory} and \ref{fig:advancedif}.
Finally, the prompt governing {Streaming Memory Maintenance} (i.e., consolidation, merging, and pruning) is illustrated in Figure~\ref{fig:memory-update}.



\begin{figure*}[t]
\centering
\begin{tcolorbox}[
  enhanced,
  width=\textwidth,
  colback=gray!6,
  colframe=black!55,
  boxrule=0.6pt,
  arc=2mm,
  left=7pt,right=7pt,top=9pt,bottom=7pt,
  colbacktitle=black!65,
  coltitle=white,
  fonttitle=\bfseries,
  title=AIME Success Memory Extraction,
  attach boxed title to top left={xshift=7pt,yshift=-2mm},
  boxed title style={
    boxrule=0pt,
    arc=1mm,
    left=6pt,right=6pt,top=2pt,bottom=2pt
  },
  before upper={\ttfamily\small\setlength{\parskip}{0pt}\setlength{\parindent}{0pt}}
]
You are a Logic Path Analyzer.

TASK:\par
Given a successful solution trace, extract reusable high-level procedural memories that summarize the correct reasoning.\par

\medskip
INPUTS:\par
[Question] :\{question\}\par
[Successful Trace] : \{reasoning\}\par

\medskip
ANALYSIS LOGIC:\par
1. Identify the dominant solution pipeline used in the Trace.\par
2. Abstract it into transferable steps that generalize beyond this single problem.\par

\medskip
GENERATION INSTRUCTIONS:\par
- Produce only 1 memory.\par
- Must be transferable: each memory must state WHEN it applies (trigger condition).\par
- Prefer actionable rules over narration.\par
- Use the format: "Step 1: [Action]. Step 2: [Action]..." (Abstract the Teacher's workflow).\par

\medskip
OUTPUT FORMAT:\par
Your output must strictly follow the format shown below:\par

MEMORY 1:\par
TITLE: <short name>\par
DESCRIPTION: <one sentence summary>\par
CONTENT: Step 1: \ldots\ Step 2: \ldots\par

\end{tcolorbox}
\caption{AIME Success Memory Extraction}
\label{AIME_success_prompt}
\end{figure*}

\begin{figure*}[t]
\centering
\begin{tcolorbox}[
  enhanced,
  breakable,
  width=\textwidth,
  colback=gray!6,
  colframe=black!55,
  boxrule=0.6pt,
  arc=2mm,
  left=7pt,right=7pt,top=9pt,bottom=7pt,
  colbacktitle=black!65,
  coltitle=white,
  fonttitle=\bfseries,
  title=AIME Failure Memory Extraction (Cascade),
  attach boxed title to top left={xshift=7pt,yshift=-2mm},
  boxed title style={
    boxrule=0pt,
    arc=1mm,
    left=6pt,right=6pt,top=2pt,bottom=2pt
  },
  before upper={\ttfamily\small\setlength{\parskip}{0pt}\setlength{\parindent}{0pt}}
]
You are a Logic Path Analyzer.

TASK:\par
Compare the [Student Trace] (Failed Path) with the [Teacher Trace] (Successful Path) as paths in a decision tree.\par
Identify the First Bifurcation Point (where the student diverged) and generate abstract Memories to fix it.\par

\medskip
INPUTS:\par
[Question] :\{question\}\par
[Student Trace]: \{reasoning\}\par
[Teacher Trace]: \{reasoning\_teacher\}\par

\medskip
ANALYSIS LOGIC:\par
1. Compare Paths: Does the Student's overall methodology (Root-to-Leaf path) match the Teacher's?\par
2. Determine Error Type:\par
\ \ \ Type A (Global Strategy Failure): The Student used a completely wrong pipeline.\par
\ \ \ Type B (Local Node Failure): The Student used the correct pipeline but made a wrong decision at a specific node due to bias, laziness, false assumption or other reasons.\par

\medskip
GENERATION INSTRUCTIONS:\par
- Produce up to \{max\_items\} memories TOTAL. Avoid overlap; merge if redundant.\par
- Must be transferable: each memory must state WHEN it applies (trigger condition).\par
- Prefer actionable rules over narration.\par

If Type A (Global Failure): Generate a "Global Procedural Memory"\par
Goal: Define the solution process.\par
Format: "Step 1: [Action]. Step 2: [Action]..." (Abstract the Teacher's workflow).\par

If Type B (Local Failure): Generate a "Local Corrective Memory"\par
Goal: Fix the specific bad decision node.\par
Format: "In context [X], do not assume [Y]. Instead, perform [Z]."\par

\medskip
OUTPUT FORMAT:\par
Your output must strictly follow the format shown below:\par

MEMORY 1:\par
TITLE: <short name>\par
DESCRIPTION: <one sentence summary>\par
CONTENT: <detailed strategy>\par

MEMORY 2:\par
\ldots
\end{tcolorbox}
\caption{AIME Failure Memory Extraction (Cascade)}
\label{AIME_prompt}
\end{figure*}

\begin{figure*}[t]
\centering
\begin{tcolorbox}[
  enhanced,
  breakable,
  width=\textwidth,
  colback=gray!6,
  colframe=black!55,
  boxrule=0.6pt,
  arc=2mm,
  left=7pt,right=7pt,top=9pt,bottom=7pt,
  colbacktitle=black!65,
  coltitle=white,
  fonttitle=\bfseries,
  title=Non-verifiable Pairwise Memory Extraction,
  attach boxed title to top left={xshift=7pt,yshift=-2mm},
  boxed title style={
    boxrule=0pt,
    arc=1mm,
    left=6pt,right=6pt,top=2pt,bottom=2pt
  },
  before upper={\ttfamily\small\setlength{\parskip}{0pt}\setlength{\parindent}{0pt}}
]
You are an AI assistant.

TASK:\par
Given an input prompt and two candidate responses, extract reusable instruction-following rules by explaining why the Better response is superior to the Worse response.\par

\medskip
INPUTS:\par
[Input/Prompt]: \{context\_text\}\par
[Better response]: \{r1\}\par
[Worse response]: \{r2\}\par

\medskip
ANALYSIS LOGIC:\par
1. Identify the decisive preference axes that make the Better response superior.\par
2. Localize the key failure points in the Worse response with respect to explicit constraints in the Input.\par
3. Convert each failure axis into a transferable conditional directive: a trigger + a concrete action that would have prevented the failure.\par

\medskip
GENERATION INSTRUCTIONS:\par
- Return 1 to 3 rules total; merge overlapping rules.\par
- Each rule must be conditional on a clear trigger (WHEN it applies).\par

\medskip
OUTPUT FORMAT:\par
Your output must strictly follow the format shown below:\par

MEMORY 1:\par
TRIGGER: The specific instruction/constraint where this rule applies (e.g., 'When asked to order items by cost', 'When the user forbids numbered lists').\par
DIMENSION: The key failure axis (e.g., format adherence, constraint satisfaction, ordering, completeness, relevance, accuracy)\par
COMPARISON: An actionable, testable directive that improves compliance. It must be conditional on the trigger (e.g., 'When ordering is requested, sort items by the specified criterion before writing the list').\par

MEMORY 2:\par

\end{tcolorbox}
\caption{Non-verifiable Pairwise Memory Extraction.}
\label{fig:nonverifiable-pairwise-memory-extraction}
\end{figure*}



\begin{figure*}[t]
\centering
\begin{tcolorbox}[
  enhanced,
  breakable,
  width=\textwidth,
  colback=gray!6,
  colframe=black!55,
  boxrule=0.6pt,
  arc=2mm,
  left=7pt,right=7pt,top=9pt,bottom=7pt,
  colbacktitle=black!65,
  coltitle=white,
  fonttitle=\bfseries,
  title=AIME Solving Prompt with Retrieved Memory,
  attach boxed title to top left={xshift=7pt,yshift=-2mm},
  boxed title style={boxrule=0pt, arc=1mm, left=6pt,right=6pt,top=2pt,bottom=2pt},
  before upper={\ttfamily\small\setlength{\parskip}{0pt}\setlength{\parindent}{0pt}}
]
You are a math problem solver. Solve problems step-by-step using clear reasoning. You may be provided with relevant problem-solving strategies from past experience. Use them if they are helpful for the current problem.\par
\medskip
=== Retrieved Problem-Solving Strategies ===\par
[Strategy 1] \{title\_1\}\par
Type: \{Global Solution Flow|Local Correction\}\par
Description: \{description\_1\}\par
\{content\_1\}\par
[Strategy 2] \{title\_2\}\par
Type: \{Global Solution Flow|Local Correction\}\par
Description: \{description\_2\}\par
\{content\_2\}\par
...\par
=== End Strategies ===\par
\medskip
Problem: \{question\}\par
Please put your final answer within \texttt{\textbackslash boxed\{\}}.\par
\end{tcolorbox}
\caption{AIME solving Prompt with Retrieved Memory}
\label{fig:aime-solve-with-memory}
\end{figure*}

\begin{figure*}[t]
\centering
\begin{tcolorbox}[
  enhanced,
  breakable,
  width=\textwidth,
  colback=gray!6,
  colframe=black!55,
  boxrule=0.6pt,
  arc=2mm,
  left=7pt,right=7pt,top=9pt,bottom=7pt,
  colbacktitle=black!65,
  coltitle=white,
  fonttitle=\bfseries,
  title=AdvancedIF Solving Prompt with Retrieved Rules,
  attach boxed title to top left={xshift=7pt,yshift=-2mm},
  boxed title style={boxrule=0pt, arc=1mm, left=6pt,right=6pt,top=2pt,bottom=2pt},
  before upper={\ttfamily\small\setlength{\parskip}{0pt}\setlength{\parindent}{0pt}}
]
You are a helpful assistant. Please answer following open-ended problems.\par
If relevant, use these instruction-following rules as guidance:\par
Preference 1: \{trigger\_1\}. For \texttt{"\{dimension\_1\}"}, \{comparison\_1\}\par
Preference 2: \{trigger\_2\}. For \texttt{"\{dimension\_2\}"}, \{comparison\_2\}\par
...\par
Preference K: \{trigger\_K\}. For \texttt{"\{dimension\_K\}"}, \{comparison\_K\}\par
Problem: \{last\_user\_request\_text\}\par
\end{tcolorbox}
\label{fig:advancedif}
\caption{AdvancedIF solving Prompt with Retrieved Memory}
\label{fig:advancedif}
\end{figure*}



\begin{figure*}[t]
\centering
\begin{tcolorbox}[
  enhanced,
  breakable,
  width=\textwidth,
  colback=gray!6,
  colframe=black!55,
  boxrule=0.6pt,
  arc=2mm,
  left=7pt,right=7pt,top=9pt,bottom=7pt,
  colbacktitle=black!65,
  coltitle=white,
  fonttitle=\bfseries,
  title=Memory Update (Refinement \& Consolidation),
  attach boxed title to top left={xshift=7pt,yshift=-2mm},
  boxed title style={
    boxrule=0pt,
    arc=1mm,
    left=6pt,right=6pt,top=2pt,bottom=2pt
  },
  before upper={\ttfamily\small\setlength{\parskip}{0pt}\setlength{\parindent}{0pt}}
]
You are a Memory Refinement Expert.\par
Context: An agent successfully solved a task and generated several "New STRATEGIES" as memories. During the task, it had also retrieved several "Historical Memories" to assist its reasoning.\par
Task: You are provided with this combined subset (New STRATEGIES + Retrieved Memories). Your goal is to refine this specific pool of memories into an optimized, consolidated set of memories.\par

\medskip
Objectives for Refinement and Consolidation:\par
1. Elimination (Redundancy Check): Identify any newly generated memories or historical memories that are completely or highly similar to others, and suggest eliminating these redundant items.\par
2. Consolidation (Merging): Identify memories that can be merged (e.g., a new memory is a more specific or general form of a historical one), and combine them into a single, more comprehensive and robust knowledge unit.\par
3. Refinement (Optimization): If the phrasing or content of a memory is not general enough or is imprecise, modify its content to increase its transferability and value for future, similar problems.\par
4. Retention (Keep): Directly retain any memory that is valuable, non-redundant, and does not require merging or refinement.\par

\medskip
Your final output must be the optimal, consolidated set of strategies.\par
Note on Historical Memories: A retrieved historical memory might have been irrelevant to the current problem's solution, but this does not necessarily make it useless. Evaluate all retrieved memories based on their generalizability and potential to help solve other problems in the future.\par

\medskip
PROBLEM THAT SUCCEED:\par
\{question\}\par

\medskip
AGENT'S REASONING TRAJECTORY:\par
\{reasoning\}\par

\medskip
NEWLY EXTRACTED ORIGINAL STRATEGIES (MEMORIES):\par
\{new\_memories\}\par

\medskip
RETRIEVED HISTORICAL MEMORIES:\par
\{retrieved\_memories\}\par

\medskip
Strictly output the final refined and consolidated strategy list in this format:\par

MEMORY 1:\par
TITLE: <concise strategy name>\par
DESCRIPTION: <one sentence summary>\par
CONTENT: <detailed transferable strategy>\par

MEMORY 2:\par
\ldots
\end{tcolorbox}
\caption{Memory Update.}
\label{fig:memory-update}
\end{figure*}

\end{document}